\documentclass[final,3p,times,twocolumn]{elsarticle}

\usepackage{amssymb}
\usepackage[pagebackref,breaklinks,colorlinks]{hyperref}
\usepackage{amsmath}
\usepackage[capitalize]{cleveref}

\journal{Pattern Recognition}

\begin{document}

\begin{frontmatter}



\title{DEVICE: Depth and Visual Concepts Aware Transformer for OCR-based Image Captioning}



\author[1]{Dongsheng Xu}
\ead{2112391059@st.gxu.edu.cn}

\author[1]{Qingbao Huang \corref{cor1}}
\ead{qbhuang@gxu.edu.cn}

\author[2]{Xingmao Zhang}
\ead{20170024@gxau.edu.cn}

\author[3]{Haonan Cheng}
\ead{haonancheng@cuc.edu.cn}

\author[1,4]{Feng Shuang}
\ead{fshuang@gxu.edu.cn}

\author[5]{Yi Cai}
\ead{ycai@scut.edu.cn}

\address[1]{School of Artificial Intelligence at Guangxi University, Nanning 530004, China}
 
\address[2]{College of General Education, Guangxi Arts University, Nanning 530022, China}
\address[3]{The State Key Laboratory of Media Convergence and Communication, Communication University of China, Beijing 100024, China}
\address[4]{Guangxi Key Laboratory of Intelligent Control and Maintenance of Power Equipment, Nanning 530004, China}
\address[5]{School of Software Engineering, South China University of Technology, Guangzhou 510006, China}

\cortext[cor1]{Corresponding author}

\begin{abstract}
OCR-based image captioning is an important but under-explored task, aiming to generate descriptions containing visual objects and scene text. Recent studies have made encouraging progress, but they are still suffering from a lack of overall understanding of scenes and generating inaccurate captions. One possible reason is that current studies mainly focus on constructing the plane-level geometric relationship of scene text without depth information. This leads to insufficient scene text relational reasoning so that models may describe scene text inaccurately. The other possible reason is that existing methods fail to generate fine-grained descriptions of some visual objects. In addition, they may ignore essential visual objects, leading to the scene text belonging to these ignored objects not being utilized. To address the above issues, we propose a Depth and Visual Concepts Aware Transformer (DEVICE) for OCR-based image captinong. Concretely, to construct three-dimensional geometric relations, we introduce depth information and propose a depth-enhanced feature updating module to ameliorate OCR token features. To generate more precise and comprehensive captions, we introduce semantic features of detected visual concepts as auxiliary information, and propose a semantic-guided alignment module to improve the model's ability to utilize visual concepts. Our DEVICE is capable of comprehending scenes more comprehensively and boosting the accuracy of described visual entities. Sufficient experiments demonstrate the effectiveness of our proposed DEVICE, which outperforms state-of-the-art models on the TextCaps test set.
\end{abstract}



\begin{keyword}
image captioning, multimodal representation, monocular depth estimation, visual object concepts


\end{keyword}

\end{frontmatter}


\section{Introduction}
\begin{figure}[h]
  \centering
   \includegraphics[width=1.0\linewidth]{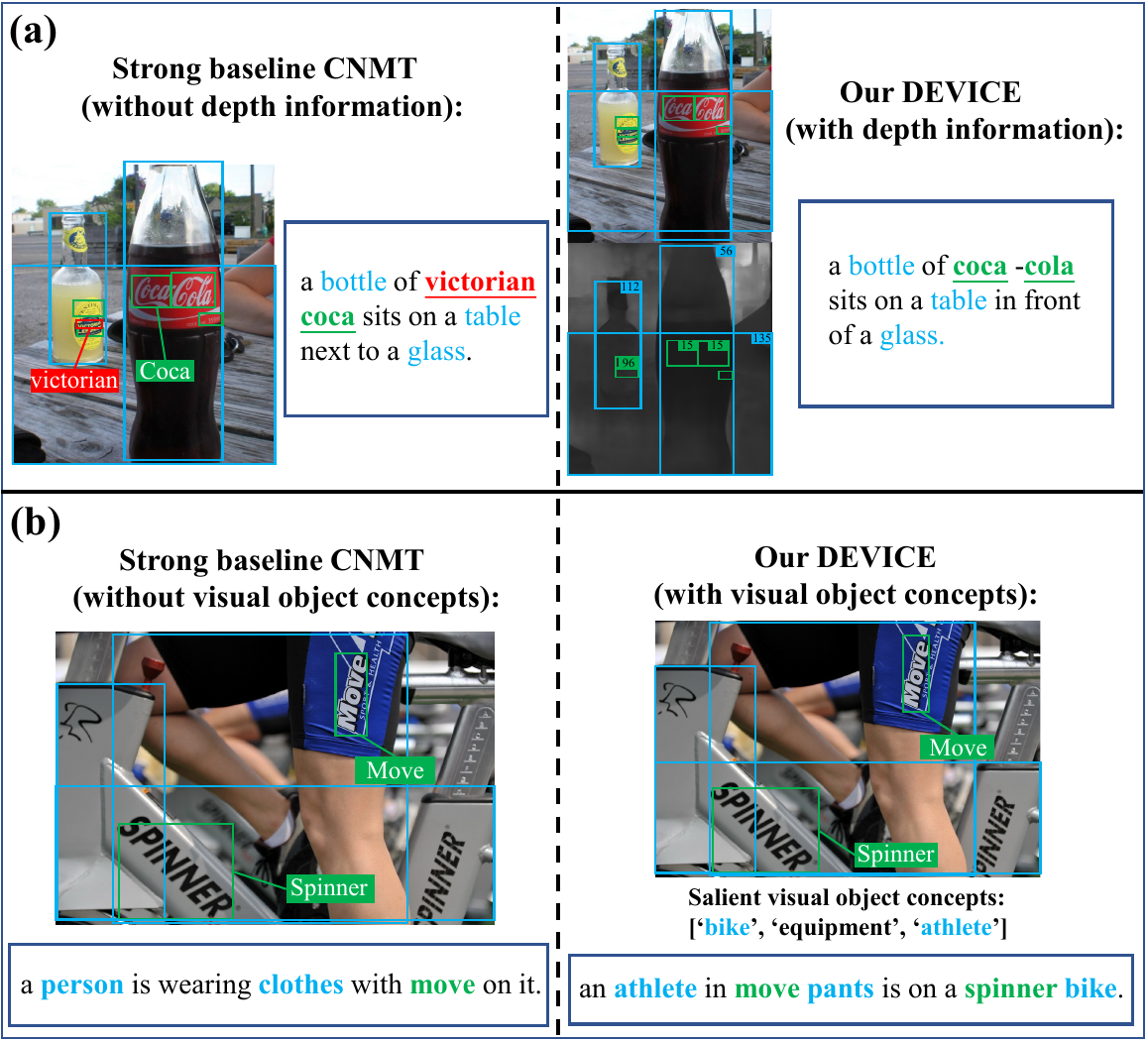}
   \caption{Our DEVICE significantly improves the quality of captions. In Case a, with depth information and a depth-enhanced feature updating module, DEVICE correctly models 3D relationships between scene texts. With a salient visual object concepts extractor and a semantic-guided alignment module, DEVICE generates more accurate and comprehensive captions, cf. Case b. Scene text is represented in \textcolor[rgb]{0,0.69,0.23}{Green}, and objects are represented in \textcolor[rgb]{0,0.76,0.98}{Blue}.}
   \label{fig:intro}
\end{figure}

Captioning is one of the important tasks in the intersection of vision and language, which aims to automatically describe given images with natural language captions. Some recent studies have achieved superior performance and are even comparable to humans under certain circumstances \cite{anderson2018bottom,lim2022protect}. However, these traditional image captioning models perform unsatisfactorily when describing scenes containing text. As shown in Case (a) (cf. \cref{fig:intro}), ``a bottle of Coca-Cola'' is obviously specific than ``a bottle''. To solve this defect, Sidorov \textit{et al.} \cite{sidorov2020textcaps} propose OCR-based image captioning and collect a high-quality dataset TextCaps, in which captions contain scene text from the~images.

To utilize the scene text in pictures, extracting Optical Character Recognition (OCR) tokens from given images is the precondition. M4C-Captioner \cite{sidorov2020textcaps} utilizes Rosetta \cite{borisyuk2018rosetta} OCR system to obtain OCR tokens, and then employ a multimodal transformer directly to model the relationships between visual objects and OCR tokens. However, it is inappropriate to treat all OCR tokens equally, because only part of OCR tokens are crucial to the understanding of the scene. Wang \textit{et al.} \cite{wang2021confidence} creatively introduce a concept of OCR confidence to choose the noteworthy OCR tokens, which has greatly alleviated this issue. Then Wang \textit{et al.} \cite{wang2020multimodal, wang2021improving} are committed to improving the spatial relationship construct methods for OCR tokens, which have been proven effective, and the proposed model LSTM-R \cite{wang2021improving} achieves the state-of-the-art performance. However, current methods are not accurate and comprehensive enough when describing the scenes, which can be mainly manifested in the following two-fold.

Firstly, the real-world is three-dimensional (3D). However, current studies \cite{sidorov2020textcaps,wang2021confidence,wang2020multimodal,wang2021improving, yang2021tap} only exploit the spatial relationships of OCR tokens at the plane-level, which leads to visual location information being utilized incompletely, and even results in generating inaccurate scene text. As illustrated in the Case (a) of \cref{fig:intro}, CNMT \cite{wang2021confidence} mistakenly regards ``victorian'' and ``coca'' as closely related OCR tokens in the same plane, which are irrelevant and actually one behind the other. Naturally, we consider that introducing depth information based on two-dimensional (2D) information can simulate real-world geometric information for improving the accuracy of scene text in captions.

Secondly, scenes contain abundant but complex visual entities (visual objects and scene text), existing methods suffer from coarse-grained descriptions of visual objects and the absence of crucial visual entities in captions. As shown in the Case (b) on the right of \cref{fig:intro}, CNMT \cite{wang2021confidence} generates the coarse-grained word ``person'' for the visual object ``athlete". Besides, CNMT ignores an important entity ``bike'', causing the OCR token ``spinner'' on it cannot be utilized effectively. One possible reason for the absence of ``bike'' is that the visual features of mutilated objects are not conducive to modeling. Intuitively, we consider that the semantic information of salient visual object concepts can alleviate coarse-grained and partial captioning.

To tackle the aforementioned issues, we propose a Depth and Visual Concepts Aware Transformer (DEVICE) for OCR-based image captioning. It mainly comprises a reading module, a depth-enhanced feature updating module, a salient visual object concepts extractor, a semantic-guided alignment module, and a multimodal transformer. Concretely, in the reading module, we employ a monocular depth estimator BTS \cite{lee2019big} to generate depth maps, then we extract depth values for visual objects and OCR tokens, respectively. The depth-enhanced feature updating module is designed to optimize the visual features of OCR tokens, which is beneficial for 3D geometry relationship construction. In the salient visual object concepts semantic extractor, we first employ a pre-trained visual-language model CLIP \cite{radford2021learning} as a filter to generate $K$ concepts most related to the given picture. Then we adopt the semantic information of visual object concepts as augmentations to improve the overall expressiveness of our model. To utilize visual concepts more efficiently, we introduce the Semantic-guided Alignment Module to interact aligned visual concepts with corresponding scene text. Subsequent experiments in \cref{sec4} demonstrate the effectiveness of our proposed DEVICE.

Our main contributions can be summarized as follows:
\begin{itemize}
\item We propose a Depth and Visual Concepts Aware Transformer for OCR-based image captinong. DEVICE performs better than previous methods as it improves the accuracy and completeness of captions.
\item We devise a Depth-enhanced Feature Updating Module (DeFUM) to enhance the 3D spatial relationship between OCR tokens, which effectively helps the model evaluate the relevance of OCR tokens. To our knowledge, we are the first to apply depth information to the OCR-based image captioning task.
\item We introduce semantic information of salient visual objects by the Visual Object Concepts Extractor (VOC), and we construct a Semantic-guided Alignment Module (SgAM) to improve the availability of visual concepts. Our DEVICE is capable of generating fine-grained objects in captions and better understanding holistic scenes.
\item Our DEVICE outperforms the state-of-the-art models on the TextCaps test set, inspiringly boosting CIDEr-D from 100.8 to 110.0.
\end{itemize}
\section{Related Work}
\subsection{OCR-based Image Captioning}
Image captioning task has made a lot of progress in recent years \cite{huang2021image,huang2019attention,li2025m3ixup}. For example, Ma \textit{et al.}\cite{li2025m3ixup} proposed a novel data augmentation method named Multi-modal Mixup to address captions that are overly simplistic and lack distinctiveness. But most of them perform poorly facing images with scene text. OCR-based image captioning was naturally born to alleviate this problem, which was proposed by Sidorov \textit{et al.} \cite{sidorov2020textcaps}. The authors introduced the Textcaps dataset and a multimodal Transformer-based \cite{vaswani2017attention} baseline model M4C-Captioner. The M4C-Captioner was modified from a Text-VQA \cite{singh2019towards,jin2022token,zeng2021beyond} model M4C \cite{hu2020iterative}, and can encode both image and OCR tokens to generate more property captions. However,  this paradigm suffers from a lack of diversity and information integrity. Zhang \textit{et al.} \cite{zhang2022magic} and Xu \textit{et al.} \cite{xu2021towards} were committed to solving this problem by introducing additional captions. Although introducing multiple captions alleviates the problem of missing scene information, it is inherently hard to summarize the complex entities of a scene within one caption well. Apart from this, M4C-Captioner is less than satisfactory performed on Textcaps due to insufficient use of visual and semantic information of images. To distinguish misidentified OCR recognition results and pick out the crucial OCR tokens, Wang \textit{et al.} \cite{wang2021confidence} introduced confidence embedding and alleviated the adverse effects of inaccurate OCR tokens to a certain extent. Yang \textit{et al.} \cite{yang2021tap} designed a pre-training model that can exploit OCR tokens and achieved strong results on multiple tasks, including OCR-based image captioning. Wang \textit{et al.} propose a scene text aware large model GIT with the aim of unifying vision-language tasks, such as image/video captioning and question answering \cite{wang2022git}. Wang \textit{et al.} \cite{wang2020multimodal,wang2021improving} and Tang \textit{et al.} \cite{tang2022ocr} employ the 2D spatial relationships to measure correlations between OCR tokens. Wang \textit{et al.} \cite{wang2023lcm} and Xu \textit{et al.} \cite{xu2023zero} focused on reducing the training overhead of models, achieving impressive results with relatively low training costs. Nevertheless, we find that simply leveraging two-dimensional spatial relationships may cause incorrect captions. Therefore, we ameliorate this problem by utilizing depth information.

\begin{figure*}[ht]
  \centering
   \includegraphics[width=0.9\linewidth]{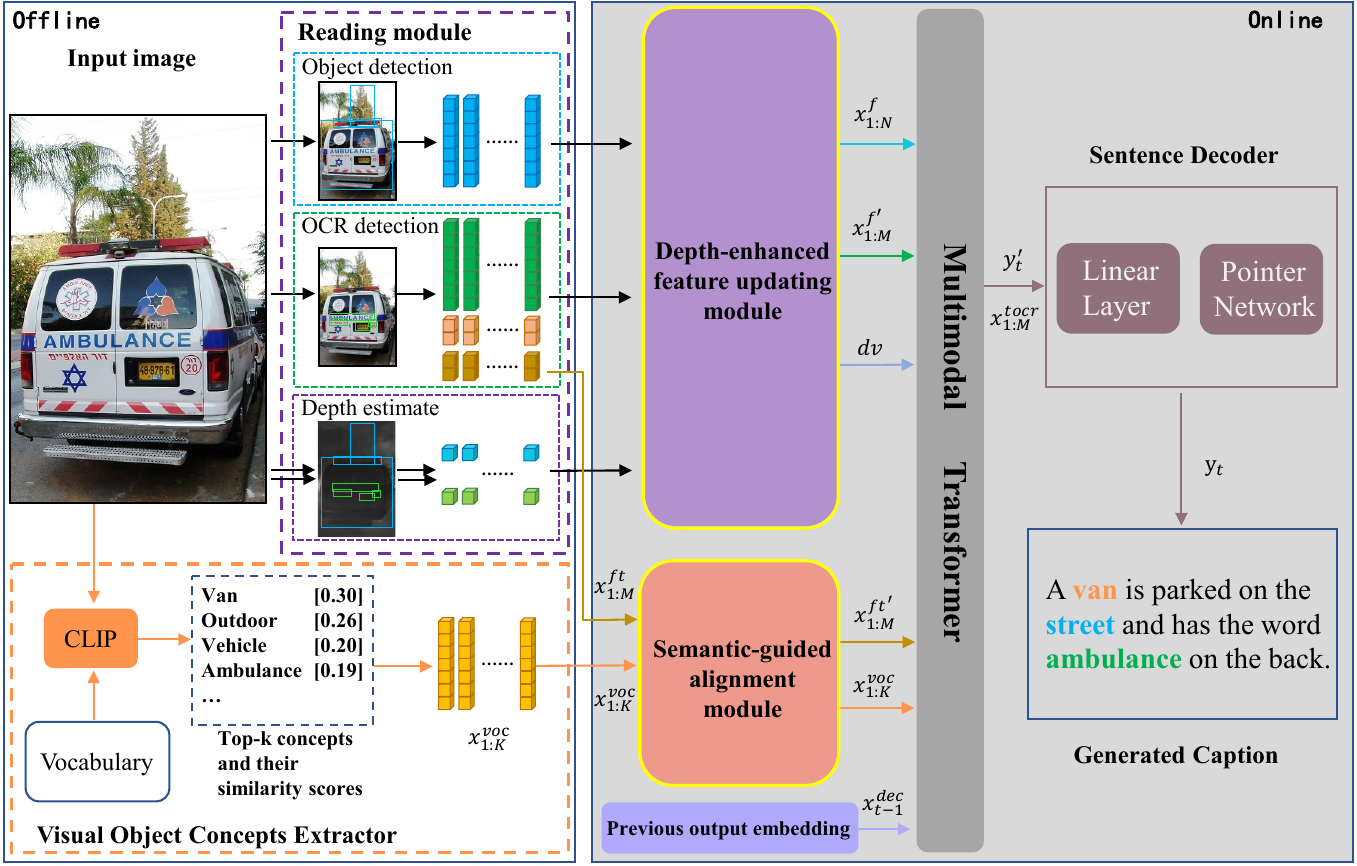}
   \caption{Overview of Depth and Visual Concepts Aware Transformer. In the reading module, we extract visual concepts along with their depth values. A depth-enhanced feature updating module is constructed to enhance OCR appearance features with the help of depth values. DEVICE utilizes depth information and plane geometry relationships to construct 3D positional relationships. Semantic information of salient visual object concepts promotes fine-grained and holistic captioning, and the proposed semantic-guided alignment module improves the model's efficiency to utilize visual concepts. Words in \textcolor[rgb]{0,0.76,0.98}{Blue}, \textcolor[rgb]{0,0.75,0.23}{Green} and \textcolor[rgb]{0.95,0.65,0}{Orange} represent objects, scene text, and visual concepts, respectively.}
   \label{fig:model}
\end{figure*}

\subsection{Depth Information For Multimodal Tasks}
Monocular depth estimation is a challenging problem of computer vision, whose purpose is to use depth values to measure the distance of objects relative to the camera \cite{lee2019big, li2023learning}. With depth estimation technology, significant improvements have been achieved on many multimodal tasks \cite{banerjee2021weakly, liao2021scene}. In the visual question answer (VQA) task, Banerjee \textit{et al.} \cite{banerjee2021weakly} and Liu \textit{et al.} \cite{9830073} proved that exploiting three-dimensional relationships by depth information could enhance the spatial reasoning ability of VQA models. Apart from this, Qiu \textit{et al.} \cite{qiu20203d} and Liao \textit{et al.} \cite{liao2021scene} point out that ignoring the depth information could tremendously place restrictions on applications of scene change captioning in the real world. At the same time, we notice that OCR-based image captioning is facing the same problem. Therefore, we improve the 3D relationship modeling method to apply in OCR-based image captioning effectively.

\section{Approach}

\subsection{Overview}

Given an image \textit{I}, OCR-based image captioning models aim to automatically generate caption $C=[y_0, y_1, ... , y_n]$ based on scene text \textit{S} in the image.  

As depicted in \cref{fig:model}, our DEVICE mainly consists of a reading module, a depth-enhanced feature updating module, a salient visual object concepts extractor, a semantic-guided alignment module, a multimodal transformer, and a sentence decoder. Concretely, the input image is first put into an object detector and two OCR detectors to extract \textit{N} visual objects and \textit{M} OCR tokens along with their 2D regional coordinates. Then we utilize a pre-trained depth estimation model to generate the pixel-level depth map \textit{D} and construct 3D coordinates of visual entities. After that, we update the visual features of OCR tokens with the depth-enhanced feature updating module. Depth information is included in the coordinates of both scene text and objects, hence the multimodal transformer could establish 3D relations between visual entities. Meanwhile, to generate salient visual object concepts for each image, we employ a pre-trained cross-modal tool CLIP \cite{radford2021learning} to match the top-\textit{K} related visual concepts (e.g., ``Van'' and ``Vehicle'') in vocabulary. Then we propose the semantic-guided alignment module to integrate the information of aligned visual concepts into the representation of relevant OCR tokens. Ultimately, the multimodal transformer and sentence decoder generate captions in the auto-regressive paradigm.

\subsection{Multimodal Embedding}
First of all, we exploit pre-trained Faster R-CNN \cite{ren2015faster} model to get the bounding box coordinates $\{b^{obj}_n\}_{n=1:N}$ of objects $\{a^{obj}_n\}_{n=1:N}$. To ensure high quality OCR tokens, following LSTM-R \cite{wang2021improving}, we apply external OCR systems \cite{borisyuk2018rosetta,Googleocr} to obtain both OCR tokens $\{a^{ocr}_m\}_{m=1:M}$ and corresponding bounding box coordinates $\{b^{ocr}_m\}_{m=1:M}$, respectively. After that, we extract the $d$-dim appearance features $\{x^{f}_n\}_{n=1:N}$ of objects and appearance features $\{x^{f}_m\}_{m=1:M}$ of OCR tokens by Faster R-CNN \cite{ren2015faster}. To generate depth maps of images, we utilize a monocular depth estimation model BTS \cite{lee2019big}.

\begin{figure}[t]
  \centering
   \includegraphics[width=1.0\linewidth]{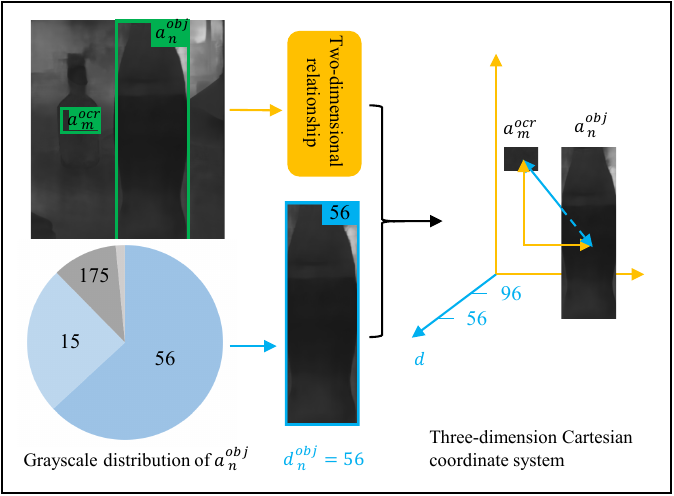}
   \caption{Illustrations for the process of getting the depth values and 3D geometric relationship modeling. The pie chart represents the gray value distribution of object $a^{obj}_n$. We take the gray values with the highest proportion as the depth value (0 means nearest) of the object $a^{obj}_n$ and OCR token $a^{ocr}_m$. $d^{obj}_n$ and $d^{ocr}_m$ denote the depth values of $a^{obj}_n$ and $a^{ocr}_m$, respectively. With depth information, relationships between visual entities become more clear. }
   \label{fig:depth}
\end{figure}

\textbf{Depth values.}
To measure the relative distance of objects and OCR tokens to the observer, we employ a pre-trained depth estimation model BTS \cite{lee2019big} to generate depth maps. Subsequently, we unify the values of the depth maps to the range of 0 to 255. For object $a^{obj}_n$ and OCR token $a^{ocr}_m$, we count the grayscale distribution of bounding box region of $a^{obj}_n$ and $a^{ocr}_m$ in the depth map at pixel scale. We take the gray values with the highest frequency as the depth values for object $a^{obj}_n$ and OCR token $a^{ocr}_m$ (cf. \cref{fig:depth}). We denote $\{dv^{obj}_n\}_{n=1:N}$ as the depth values for $\{a^{obj}_n\}_{n=1:N}$ and $\{dv^{ocr}_m\}_{m=1:M}$ for $\{a^{ocr}_m\}_{m=1:M}$, respectively.

\textbf{Embedding of objects.}
To measure the relationship between visual entities more effectively, we incorporate depth information into object features. For the object $a^{obj}_n$, we denote its 3D geometry coordinate by $b^{obj}_n = [x_{min}/W; y_{min}/H; x_{max}/W; y_{max}/H; dv^{obj}_n/255]$, where $W$ and $H$ represent the width and height of the image $I$, respectively. The embedding of object $a^{obj}_n$ is calculated as
\begin{equation}
    x^{obj}_n = LN(W_{f1}x^{f}_n) + LN(W_{b1}b^{obj}_n),
\end{equation}
where $W_{f1} \in \mathbb{R}^{t\times{d}}$ and $W_{b1} \in \mathbb{R}^{t\times{5}}$, $t$ represents the~dimension of common space in multimodal transformer, and $LN$ represents layer normalization.

\textbf{Embedding of OCR tokens.}
Unlike objects, OCR tokens not only contain rich visual clues but also contain rich semantic information. To get rich representations of OCR tokens, following Hu \textit{et al.} \cite{hu2020iterative} and Sidorov \textit{et al.} \cite{sidorov2020textcaps}, we apply FastText \cite{bojanowski2017enriching} to extract the 300-dim sub-word feature $x^{ft}_m$ for OCR token $a^{ocr}_m$, and employ PHOC (Pyramidal Histogram of Characters) \cite{almazan2014word} to extract 604-dim character-level feature $x^{ph}_m$ for $a^{ocr}_m$. Similarly, the spatial location of $a^{ocr}_m$ is denoted as $b^{ocr}_m = [x_{min}/W; y_{min}/H; x_{max}/W; y_{max}/H; dv^{ocr}_m/255]$. To better fuse the three-dimensional relative spatial relationships into OCR features, we propose a Depth-enhanced Feature Updating Module (DeFUM, cf. \cref{subsec c}). In the real world, scene texts are attached to the surface of objects. We consider that the visual features of objects can help model the visual relationship of OCR tokens. For example, if several different OCR tokens are on the same object, then these OCR tokens are likely to have a strong correlation. Therefore we fed the OCR token appearance features $x^{f}_{1:M}$ into DeFUM along with object appearance features $x^{f}_{1:N}$, we can get the depth-enhanced OCR appearance feature $x^{f'}_{1:M}$: 
\begin{equation}
    x^{f'}_{1:M} = DeFUM(x^{f}_{1:N},x^{f}_{1:M},dv^{obj}_{1:N},dv^{ocr}_{1:M}),
\end{equation}
where $dv^{obj}_{1:N} \in \mathbb{R}^{N\times{1}}$ and $dv^{ocr}_{1:M} \in \mathbb{R}^{M\times{1}}$, respectively.
Considering the gap between visual features and semantic features, we do not update FastText and PHOC features (semantic embeddings).
Then we utilize the Semantic-guided Alignment Module (SgAM, cf. \cref{subsec e}) to integrate aligned visual concept information into the semantic representation of relevant OCR tokens
\begin{equation}
    x^{ft'}_{1:M} = SgAM(FT(a^{voc}_{1:K}),x^{ft}_{1:M}),
\end{equation}
where $x^{ft'}_{1:M}$ and $x^{ft}_{1:M}\in \mathbb{R}^{M\times{300}}$, $FT(a^{voc}_k)$ represent the FastText embedding of visual concept $a^{voc}_k$ (cf. \cref{subsec d}).

Following Wang \textit{et al.} \cite{wang2021confidence}, we adopt confidence features of detected OCR tokens $\{x^{conf}_m\}_{m=1:M}$ due to some OCR tokens may be amiss. The final embedding of OCR token $a^{ocr}_m$ could be calculated by
\begin{equation}
\begin{aligned}
    x^{ocr}_m = LN(W_{f2}x^{f'}_m + W_{ft}x^{ft'}_m + W_{ph}x^{ph}_m)\\ + LN(W_{b2}b^{ocr}_m) + LN(W_{conf}x^{conf}_m),
\end{aligned}
\end{equation}
where $W_{f2} \in \mathbb{R}^{t\times{d}}$, $W_{ft} \in \mathbb{R}^{t\times{300}}$,  $W_{ph} \in \mathbb{R}^{t\times{604}}$, $W_{b2} \in \mathbb{R}^{t\times{5}}$, and $W_{conf} \in \mathbb{R}^{t\times{1}}$ are learnable parameters, and $LN$ represents layer normalization.

\begin{figure}[t]
  \centering
   \includegraphics[width=1.0\linewidth]{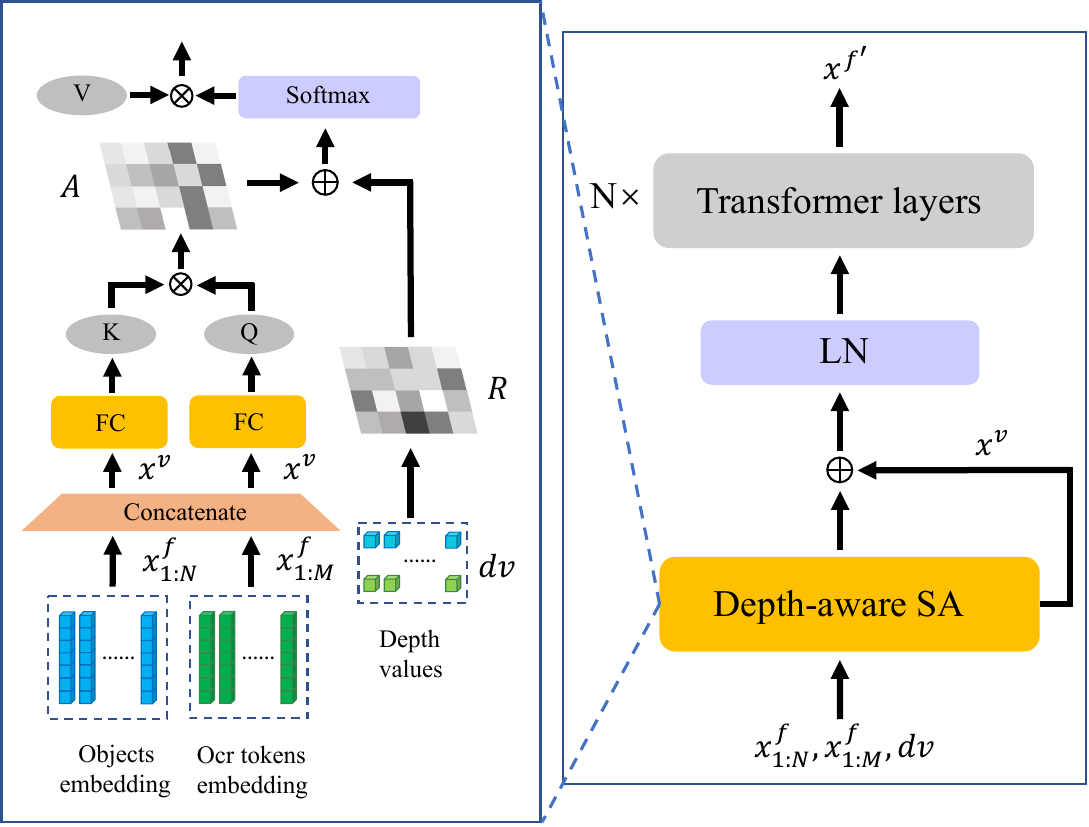}
   \caption{Illustration of the Depth-enhanced Feature Updating Module (DeFUM), which consists of a depth-aware self-attention module and $N$ layers of transformer encoder. The DeFUM could effectively enhance the appearance features of OCR tokens with the help of depth information, which is conducive to constructing the three-dimensional geometric relationships and improving the accuracy of captions.}
   \label{fig:dem}
\end{figure}
\subsection{Depth-enhanced Feature Updating Module}
\label{subsec c}
Unlike tasks like VQA, which need high-quality object relationships, OCR-based image captioning demands strong OCR relationship modeling capabilities. Thus, the quality of OCR features is crucial for captions. For this reason, we propose DeFUM to enhance the OCR appearance features $\{x^{f}_m\}_{m=1:M}$ with the help of depth information. Intuitively, we consider that object features can play the role of a bridge linking its adjacent scene text. Therefore, we first obtain appearance embedding $x^v$ of visual entities by concatenating the visual features $\{x^{f}_n\}_{n=1:N}$ of objects $\{a^{obj}_n\}_{n=1:N}$ and the visual features $\{x^{f}_m\}_{m=1:M}$ of OCR tokens $\{a^{ocr}_m\}_{m=1:M}$: 
\begin{equation}
    x^v = Concat(x^{f}_{1:N}, x^{f}_{1:M}),
\end{equation}
where $x^v \in \mathbb{R}^{(m+n)\times{d}}$. Subsequently, we calculate the query $Q$, key $K$, and value $V$ as follows:
\begin{equation}
    \left\{ 
    \begin{aligned}
    Q &= x^{v}W_Q \\
    K &= x^{v}W_K,\\
    V &= x^{v}W_V \\
    \end{aligned} 
    \right.
\end{equation}
where query $Q$, key $K$ and value $V \in \mathbb{R}^{(m+n)\times{d}}$. Then, we can obtain the attention weights matrix $A$ by: 
\begin{equation}
    A = \frac{QK^T}{\sqrt{d}},
\end{equation}
where $K \in \mathbb{R}^{(m+n)\times{(m+n)}}$. To measure the relative depth between different visual entities, we calculate the relative depth score of $x^v_i$ and $x^v_j$ by
\begin{equation}
    R_{i,j} = log(\frac{min(dv_j, dv_i)}{max(dv_j, dv_i)}),
\end{equation}
where $dv = Concat(dv^{obj},dv^{ocr})$ and the relative depth map $R\in \mathbb{R}^{(m+n)\times{(m+n)}}$. Then the output of the Depth-aware Self-Attention module (cf. \cref{fig:dem}) is updated by 
\begin{equation}
    x^{f'} = transformer(LN(x^v + softmax(A + R)V)),
\end{equation}
where the depth-aware visual feature $x^{f'}\in \mathbb{R}^{m\times{d}}$ and $LN$ denotes layer normalization. With the help of $R$, the attention between visually entities with significant differences in relative depth is weakened, while the entities with smaller differences in relative depth are less affected. In the Transformer layers, self-attention mechanism provides better modelling ability for inter- and intra-modal data than other attention-based mechanisms \cite{wan2022revisiting}.

\subsection{Salient Visual Object Concepts Extractor}
\label{subsec d}
To improve the modeling ability of the multimodal transformer to the scene objects, avoid the lack of vital objects, and generate more accurate captions, we propose a visual object concepts extractor to introduce semantic information of visual object concepts. Specifically, we employ a pre-trained model CLIP \cite{radford2021learning} to match $15$ object concepts related to the given image from vocabulary. Then we sort them by similarity score and take the top-K as visual object concepts. Visual object concepts and similarity score are represented as $\{a^{voc}_k\}_{k=1:K}$ and $\{a^{score}_k\}_{k=1:K}$. The embedding of visual object concepts are calculated as follows:
\begin{equation}
     x^{voc}_k = LN(W_{voc}FT(a^{voc}_k))\\ + LN(W_{score}a^{score}_k),
\end{equation}
where $FT(a^{voc}_k)$ means the FastText \cite{bojanowski2017enriching} embedding of $a^{voc}_k$, $W_{voc} \in \mathbb{R}^{t\times{300}}$ and $W_{score} \in \mathbb{R}^{t\times{1}}$.

\subsection{Semantic-guided Alignment Module}
\label{subsec e}
CLIP \cite{radford2021learning} is a large-scale vision-language model employing contrastive learning, proven effective in downstream tasks \cite{zhang2022pointclip}. However, CLIP embeds entire images and sentences, hindering its ability to capture fine-grained features \cite{ma2022x}. This makes it challenging to prioritize important visual concepts, potentially harming model performance and convergence efficiency.
\begin{figure}[t]
  \centering
   \includegraphics[width=1.0\linewidth]{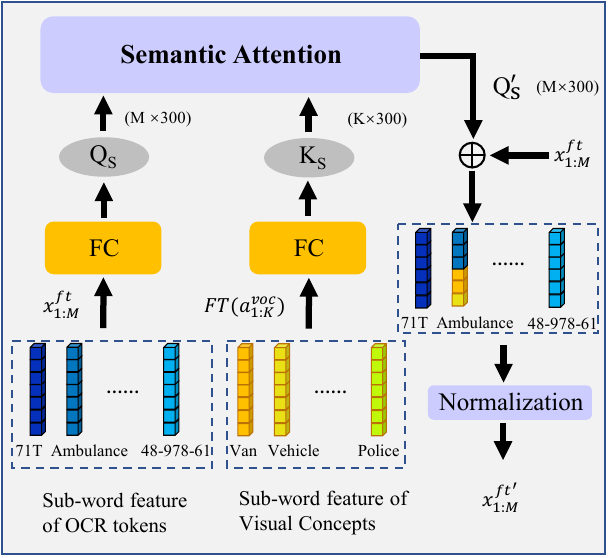}
   \caption{Illustration of the Semantic-guided Alignment Module (SgAM), which utilizes semantic attention to transfer the information of visual concepts to semantically aligned OCR Tokens' sub-word embedding. This operation is capable of making model emphasize visual concepts that are highly relevant to the scene text. Besides, SgAM helps multimodal transformer roughly model the spatial location of semantically aligned visual concepts.}
   \label{fig:sgam}
\end{figure}
To address this issue, we propose a Semantic-guided Alignment Module (SgAM, cf. \cref{fig:sgam}). By leveraging FastText trained on Wikipedia, which captures semantic similarities (e.g., "Van" and "Vehicle"), we integrate semantic information of visual concepts into related OCR token features. This aids in distinguishing some relevant visual concepts and models their spatial positions during interaction with OCR tokens in the multimodal transformer.

\begin{table*}
  \centering
    \caption{Comparison of our performance with other baseline models on the TextCaps validation set. TAP$^{\dagger}$ and ConCap$^{\dagger}$ are pre-trained models. Our DEVICE outperforms state-of-the-art models on METOR, ROUGE-L, SPICE, and CIDEr-D.}
    \label{tab1}
  \resizebox{2.0\columnwidth}{!}{
  \begin{tabular}{l|ccccc}
    \hline
    Model & BLEU-4 & METEOR & ROUGE-L & SPICE & CIDEr-D \\
    \hline
    \hline
    Up-Down (CVPR 2018) \cite{anderson2018bottom} & 20.1 & 17.8 & 42.9 & 11.7 & 41.9 \\
    AoA (ICCV 2019) \cite{huang2019attention} & 20.4 & 18.9 & 42.9 & 13.2 & 42.7 \\
    M4C-Captioner (ECCV 2020) \cite{sidorov2020textcaps} & 23.3 & 22.0 & 46.2 & 15.6 & 89.6 \\
    MMA-SR (ACM MM 2020) \cite{wang2020multimodal} & 24.6 & 23.0 & 47.3 & 16.2 & 98.0\\
    SS-Baseline (AAAI 2021) \cite{zhu2021simple} & 24.9 & 22.7 & 47.2 & 15.7 & 98.8\\
    CNMT (AAAI 2021) \cite{wang2021confidence} & 24.8 & 23.0 & 47.1 & 16.3 & 101.7\\
    ACGs-Captioner (CVPR 2021) \cite{xu2021towards} & 24.7 & 22.5 & 47.1 & 15.9 & 95.5\\
    MAGIC (AAAI 2022) \cite{zhang2022magic} & 22.2 & 20.5 & 42.3 & 13.8 & 76.6\\
    LCM (Neural Networks 2023) \cite{wang2023lcm} & 23.9 & 22.5 & 46.6 & 16.1 & 92.4\\
    Zero-TextCap (ACM MM 2024) \cite{xu2023zero} & 2.6 & 10.6 & 21.7 & 4.6 & 22.9\\
    OMOT (ICMR 2022) \cite{tang2022ocr} & 26.4 & 22.2 & 47.5 & 14.9 & 95.1\\
    LSTM-R (CVPR 2021) \cite{wang2021improving} & \textbf{27.9} & 23.7 & 49.1 & 16.6 & 109.3\\
    CFGR (Neurocomputing 2024) \cite{zhou2024cross} & 25.6 & 23.3 & 47.7 & 16.5 & 105.7\\
    \hline
    DEVICE (Ours) & 27.6 & \textbf{24.6} & \textbf{49.3} & \textbf{17.7} & \textbf{117.1}\\
    \hline
    M4C-Captioner(w/ GT OCRs) \cite{sidorov2020textcaps} & \textcolor[rgb]{0.5,0.5,0.5}{26.0} & \textcolor[rgb]{0.5,0.5,0.5}{23.2} & \textcolor[rgb]{0.5,0.5,0.5}{47.8} & \textcolor[rgb]{0.5,0.5,0.5}{16.2} & \textcolor[rgb]{0.5,0.5,0.5}{104.3}\\
    TAP$^{\dagger}$ (CVPR 2021) \cite{yang2021tap} & 25.8 & 23.8 & 47.9 & 17.1 & 109.2\\
    ConCap$^{\dagger}$ (AAAI 2023) \cite{wang2022controllable} & 31.3 & - & - & - & 116.7\\
    \hline
  \end{tabular}
  }
\end{table*}

Concretely, we first convert FastText embedding of visual concepts $FT(a^{voc}_{1:K})$ to query $K_S$ and convert FastText embedding of OCR tokens $x^{ft}_{1:M}$ to value $Q_S$, respectively:
\begin{equation}
    \left\{ 
    \begin{aligned}
    Q_S &=  x^{ft}_{1:M}W_{Q_S}\\
    K_S &= FT(a^{voc}_{1:K})W_{K_S},\\
    \end{aligned} 
    \right.
\end{equation}
where $FT(a^{voc}_{1:K})\in \mathbb{R}^{K\times{300}}$, $x^{ft}_{1:M}\in \mathbb{R}^{M\times{300}}$, while $W_{Q_S}$ and $W_{K_S}\in \mathbb{R}^{300\times{300}}$.

Subsequently, we adopt semantic attention to interact semantic information of visual concepts and OCR tokens
\begin{equation}
     Q'_S = softmax(\frac{Q_SK_S^T}{\sqrt{d_s}})^TFT(a^{voc}_{1:K}),
\end{equation}
where $Q'_S\in \mathbb{R}^{M\times{300}}$. Then we can get the final semantic embedding $x^{ft'}_{1:M}$ of OCR tokens
\begin{equation}
     x^{ft'}_{1:M} = L2Norm(x^{ft}_{1:M} + Q'_S),
\end{equation}
where $x^{ft'}_{1:M}$ and $x^{ft}_{1:M}\in \mathbb{R}^{M\times{300}}$, while $L2Norm$ represents L2 Normalization.

\subsection{Reasoning and Generation Module}
Following Sidorov \textit{et al.} \cite{sidorov2020textcaps}, we adopt a multimodal transformer to encode all the obtained embeddings. Since many OCR tokens (e.g., “coca” and “spinner” in \cref{fig:intro}) are uncommon words, generating captions from a fixed vocabulary is unsuitable. Following M4C-Captioner \cite{sidorov2020textcaps}, we utilize two classifiers for common vocabulary and candidate OCR tokens separately to generate words. Overall, multimodal transformer models multimodal embedding and generates common words $y'_t$. Afterward, a dynamic pointer network \cite{vinyals2015pointer} is employed to generate final words $y_t$:
\begin{equation}
\begin{aligned}
     y'_t, x^{tocr}_{1:M} = mmt(x^{obj}_{1:N}, x^{ocr}_{1:M}, x^{ft'}_{1:M}, x^{voc}_{1:K}, x^{dec}_{t-1}),
\end{aligned}
\end{equation}
where $mmt$ represents multimodal transformer, $x^{tocr}_m$ indicates the feature of $x^{ocr}_m$ updated by $mmt$, $x^{ft'}_{1:M}$ represents semantic
embedding updated by SgAM, and $x^{dec}_{t-1}$ is the embedding of previous output $y'_{t-1}$, respectively. Eventually, the dynamic pointer network predicts using both common words from the fixed vocabulary and special words $x^{tocr}_{1:m}$ from OCR tokens:
\begin{equation}
     y_t = argmax([l_{m}^{w}(y'_t),PN(y'_t, x^{tocr}_{1:m})]),
\end{equation}
where $l_{m}^{w}(\cdot)$ denotes a linear classifier for vocabulary, $PN(\cdot)$ denotes pointer network. We train our model by optimizing the Cross-Entropy loss:
\begin{equation}
     L(\theta) = -\sum_{t=1}^{T}log(\hat{y}_t|y_{1:t-1}, x^{tocr}_{1:m}),
\end{equation}
where $\hat{y}_t$ is the corresponding token in the ground truth.

\section{Experiments}
\label{sec4}

\begin{table*}
  \centering
    \caption{Comparison of our performance with other baseline models on the TextCaps test set. Our DEVICE outperforms state-of-the-art models on all metrics, notably boosting CIDEr-D from 100.8 to 110.0.}
    \label{tab2}
  \resizebox{2.0\columnwidth}{!}{
  \begin{tabular}{l|ccccc}
    \hline
    Model & BLEU-4 & METEOR & ROUGE-L & SPICE & CIDEr-D \\
    \hline
    \hline
    Up-Down (CVPR 2018) \cite{anderson2018bottom} & 14.9 & 15.2 & 39.9 & 8.8 & 33.8 \\
    AoA (ICCV 2019) \cite{huang2019attention} & 15.9 & 16.6 & 40.4 & 10.5 & 34.6 \\
    M4C-Captioner (ECCV 2020) \cite{sidorov2020textcaps} & 18.9 & 19.8 & 43.2 & 12.8 & 81.0 \\
    MMA-SR (ACM MM 2020) \cite{wang2020multimodal} & 19.8 & 20.6 & 44.0 & 13.2 & 88.0\\
    SS-Baseline (AAAI 2021) \cite{zhu2021simple} & 20.2 & 20.3 & 44.2 & 12.8 & 89.6\\
    CNMT (AAAI 2021) \cite{wang2021confidence} & 20.0 & 20.8 & 44.4 & 13.4 & 93.0\\
    ACGs-Captioner (CVPR 2021) \cite{xu2021towards} & 20.7 & 20.7 & 44.6 & 13.4 & 87.4\\
    LCM (Neural Networks 2023) \cite{wang2023lcm} & 19.3 & 20.5 & 43.8 & 13.2 & 83.1\\
    Zero-TextCap (ACM MM 2024) \cite{xu2023zero} & 2.3 & 10.6 & 21.9 & 4.5 & 24.0\\
    OMOT (ICMR 2022) \cite{tang2022ocr} & 21.2 & 19.6 & 44.4 & 12.1 & 84.9\\
    LSTM-R (CVPR 2021) \cite{wang2021improving} & 22.9 & 21.3 & 46.1 & 13.8 & 100.8\\
    CFGR (Neurocomputing 2024) \cite{zhou2024cross} & 20.9 & 21.3 & 45.1 & 13.7 & 98.2\\
    \hline
    DEVICE (Ours) & \textbf{23.1} & \textbf{22.5} & \textbf{46.7} & \textbf{15.0} & \textbf{110.0}\\
    \hline
    M4C-Captioner(w/ GT OCRs) \cite{sidorov2020textcaps} & \textcolor[rgb]{0.5,0.5,0.5}{21.3} & \textcolor[rgb]{0.5,0.5,0.5}{21.1} & \textcolor[rgb]{0.5,0.5,0.5}{45.0} & \textcolor[rgb]{0.5,0.5,0.5}{13.5} & \textcolor[rgb]{0.5,0.5,0.5}{97.2}\\
    TAP$^{\dagger}$ (CVPR 2021) \cite{yang2021tap} & 21.9 & 21.8 & 45.6 & 14.6 & 103.2\\
    ConCap$^{\dagger}$ (AAAI 2023) \cite{wang2022controllable} & 27.4 & - & - & - & 105.6\\
    Human \cite{sidorov2020textcaps} & 24.4 & 26.1 & 47.0 & 18.8 & 125.5\\
    \hline
  \end{tabular}
  }
\end{table*}

\subsection{Dataset and Settings}
\textbf{Dataset and evaluation metrics.}
The TextCaps dataset \cite{sidorov2020textcaps} is constructed for OCR-based image captioning, which contains 28408 images with 5 captions per image.  The training, validation, and test set contain 21953, 3166, and 3289 images, respectively. We choose five widely used evaluation metrics BLEU-4 \cite{papineni2002bleu}, ROUGE-L \cite{liao2017textboxes}, METEOR  \cite{banerjee2005meteor}, SPICE \cite{anderson2016spice}, and CIDEr-D  \cite{vedantam2015cider} for evaluation. Following Sidorov \textit{et al.} \cite{sidorov2020textcaps}, we focus on CIDEr-D when comparing different methods. CIDEr-D strongly correlates with human evaluation and emphasizes informative special tokens like OCR tokens.

\textbf{Settings and implementation details.}
To enhance OCR system robustness and detection accuracy, we utilize the Rosetta OCR \cite{borisyuk2018rosetta} and Google OCR \cite{Googleocr}, following Wang \textit{et al.} \cite{wang2021improving}. Each image is restricted to a maximum of 80 OCR tokens. We extract 100 object appearance features per image, with feature dimension $d=2048$. Common embedding space dimension $t=768$ in the multimodal transformer. The maximum generation length is 30. DeFUM employs a two-layer transformer encoder; the multimodal transformer has 4 layers and 12 self-attention heads, following BERT-BASE \cite{devlin2018bert} default settings. The vocabulary contains 6,736 words. To get visual object concepts, we filter top-$5$ visual concepts from the vocabulary by CLIP \cite{radford2021learning}. We retain CLIP's similarity score to assess concept relevance. For OCR token confidence, Rosetta-detected tokens use Wang \textit{et al.} \cite{wang2021confidence} metrics. Tokens unique to one OCR system are assigned a confidence of $0.9$.

We train the model for about 26 epochs (9000 iterations) on a single 3090 Ti GPU, and the batch size is 64. We adapt Adam \cite{kingad2015} optimizer, the initial learning rate is $1e$-$4$ and is declined to 0.1 times at 7000 iterations. We monitored the CIDEr-D metric to choose the best model and evaluate it on both the validation set and test set. All the experimental results are computed by Eval AI online platform submissions.

\subsection{Experimental Results}
\textbf{Compared models.}
We make comparisons with other models in \cref{tab1} and \cref{tab2}. \textbf{M4C-Captioner} \cite{sidorov2020textcaps}, \textbf{SS-Baseline} \cite{zhu2021simple}, \textbf{CNMT} \cite{wang2021confidence}, and \textbf{ACGs-Captioner} \cite{xu2021towards} are Transformer-based strong baseline models. \textbf{MMA-SR}  \cite{wang2020multimodal}, \textbf{MAGIC} \cite{zhang2022magic}, \textbf{OMOT}   \cite{tang2022ocr}, and \textbf{LSTM-R} \cite{wang2021improving} are LSTM-based strong baseline models. LCM \cite{wang2023lcm} is a lightweight model and Zero-TextCap \cite{xu2023zero} is a zero-shot method. 
\textbf{Main results on the TextCaps validation set.}
The comparisons on the validation set between our DEVICE and other models are shown in \cref{tab1}. Traditional methods UP-Down and AOA perform less well than other models with OCR tokens due to they cannot generate scene text. CNMT proves the effectiveness of confidence embedding and repetition mask, which improves CIDEr-D from $89.6$ to $101.7$ compared with M4C-Captioner. LSTM-R constructs sufficient 2D spatial relations and boosts all metrics significantly. By incorporating depth information and semantic information of visual object concepts with transformer, compared with SOTA model LSTM-R \cite{wang2021improving}, our DEVICE improves the performance on METEOR, SPICE, CIDEr-D by $0.9$, $1.1$, and $7.8$, respectively. The performance of DEVICE on BLEU-4 ranks second only to LSTM-R (0.3 below). Note that LSTM-R has cleaned the training captions by removing the undetected OCR symbols, which increases BLEU-4 by 0.6 \cite{wang2021improving}. For a fair comparison, we train our model with raw captions provided by TextCaps \cite{sidorov2020textcaps}, following the majority \cite{sidorov2020textcaps, wang2021confidence, zhu2021simple, yang2021tap}. Moreover, CIDEr-D and METEOR have the highest correlation with human scores \cite{sidorov2020textcaps}, which shows that DEVICE significantly outperforms other baseline models from a human perspective. The performance of our model on the TextCaps validation set demonstrates the effectiveness of our DEVICE.
\begin{table}[t]
  \centering
  \caption{Comparison of iterations, time and computational complexity (FLOPs) of DEVICE and other baseline models required to converge to the optimal CIDER. When computiong FLOPs, we omit all vector dimensions $d$ for simplicity.}
  \label{tab4}
  \resizebox{1.0\columnwidth}{!}{
  \begin{tabular}{l|ccccc}
    \hline
    Model & Batch Size & Iterations/epochs & Time & FLOPs (Encoder)\\
    \hline
    \hline
    M4C-Captioner & 128 & 12000/70 & 11h & 176400\\
    CNMT & 128 & 12000/70 & 12h & 176400 \\
    ACGs-Captioner & 128 & 12000/70 & 12h & 362800\\
    SS-Baseline & 128 & 15000/87 & 14h & 59168 \\
    LSTM-R & 50 & 12956/30 & - & -\\
    \hline
    DEVICE(w/o SgAM) & 64 & 12000/35 & 11h & 241200\\
    DEVICE & 64 & 9000/26 & 8h & 265600\\
    \hline
    TAP$^{\dagger}$ & - & 480000/- & - & -\\
    ConCap$^{\dagger}$ & 2880 & -/37 & - & -\\
    \hline
  \end{tabular}
  }
\end{table}
\begin{table}[t]
  \centering
  \caption{Ablation of each module in DEVICE on the TextCaps validation set. Google denotes Google OCR system, DI denotes Depth Information, DeFUM represents the Depth-enhanced Feature Updating Module, VOC indicates semantic information of Visual Object Concepts, and SgAm represents Semantic-guided Alignment Module.}
  \label{tab3}
  \resizebox{1.0\columnwidth}{!}{
  \begin{tabular}{lccccc|cc}
    \hline
    & Google & DI & DeFUM & VOC & SgAM & BLEU-4 & CIDEr-D \\
    \hline
    \hline
    1 & & & & & & 24.5 &101.5 \\
    2 & \checkmark & & & & & 24.9 & 105.3 \\
    3 & \checkmark & \checkmark & & & & 25.6 & 109.8 \\
    4 & \checkmark & \checkmark & \checkmark & & & 26.1 & 113.9\\
    5 & \checkmark & & & \checkmark & & 26.3 & 112.4\\
    6 & \checkmark & & & \checkmark & \checkmark & 26.5 & 114.4\\
    7 & \checkmark & \checkmark & \checkmark & \checkmark & & 27.4 & 115.9\\
    8 & & \checkmark & \checkmark & \checkmark & \checkmark & 26.9 & 115.0\\
    9 & \checkmark & \checkmark & \checkmark & \checkmark & \checkmark & \textbf{27.6} & \textbf{117.1}\\
    \hline
  \end{tabular}
  }
\end{table}

\textbf{Main results on the TextCaps test set.}
Compared with LSTM-R, our model shows superiority on the TextCaps test set, which gains rises of $0.2$, $1.2$, $0.6$, $1.2$, $9.2$ on BLEU-4, METEOR, ROUGE-L, SPICE, CIDEr-D, respectively. Significantly, the inspiring improvement on the CIDEr-D shows that our DEVICE generates more accurate scene text. The results of m4c-captioner (w/ GT OCRs) \cite{sidorov2020textcaps} are evaluated on a subset of the TextCaps test set, excluding those samples without OCR annotations. DEVICE outperforms m4c-captioner (w/ GT OCRs) verifies that our DEVICE is capable of compensating the negative impact of the error of OCR systems to some extent. Encouragingly, our model outperforms the pre-trained model TAP$^{\dagger}$ on all metrics, demonstrating that DEVICE utilizes limited training data more efficiently. The BLEU-4 score of ConCap$^{\dagger}$ is 27.3, which shows that ConCap$^{\dagger}$ has learned human language ability well from the large-scale corpus. However, DEVICE outperforms ConCap$^{\dagger}$ on CIDEr-D, which shows that the scene text in the captions generated by DEVICE is of higher quality. Finally, the gap between machines and humans narrows on the OCR-based image captioning task. 

\textbf{Analysis of model convergence speed and computational complexity.}
In our experiments, we find that SgAM can significantly improve the model's converge speed, and achieve better results with fewer iterations. \cref{tab4} presents the iteration counts and training epochs required for various models to attain optimal CIDEr-D scores. For computational complexity, \cref{tab4} details the FLOPs (Floating Point Operations) of our DEVICE alongside baseline models including M4C-Captioner, CNMT, ACGs-Captioner, and SS-Baseline. Our DEVICE demonstrates significantly faster convergence speed compared to the baseline models. Due to the introduction of DeFUM and SgAM, our model exhibits slightly higher FLOPs compared to M4C-Captioner and CNMT, but significantly lower than ACGs-Captioner. Although the introduction of SgAM increases the computational complexity (241200 vs. 265600 FLOPs), the required training time for DEVICE is reduced from 11 hours to 8 hours. This proves that semantic attention can help the multimodal transformer quickly select effective visual entities. M4C-Captioner, CNMT, ACGs-Captioner, and SS-Baseline are trained on 2 RTX 3090 Ti GPUs in \cref{tab4}.

\begin{table}[t]
  \centering
  \caption{Analysis of visual object concepts number $K$ on validation set. VOC indicates Visual Object Concepts.}
  \label{tab5}
  \resizebox{1.0\columnwidth}{!}{
  \begin{tabular}{lc|ccc}
    \hline
    Model & Number of VOC & CIDEr-D \\
    \hline
    \hline
    DEVICE & $K$ = 3 & 115.9 & \textcolor{red}{$\downarrow$}1.2\\
    DEVICE & $K$ = 5 & \textbf{117.1} &\\
    DEVICE & $K$ = 8 & 116.5 & \textcolor{red}{$\downarrow$}0.6\\
    DEVICE & $K$ = 10 & 116.1 & \textcolor{red}{$\downarrow$}1.0\\
    \hline
  \end{tabular}
  }
\end{table}

\begin{figure*}[t]
  \centering
   \includegraphics[width=1\linewidth]{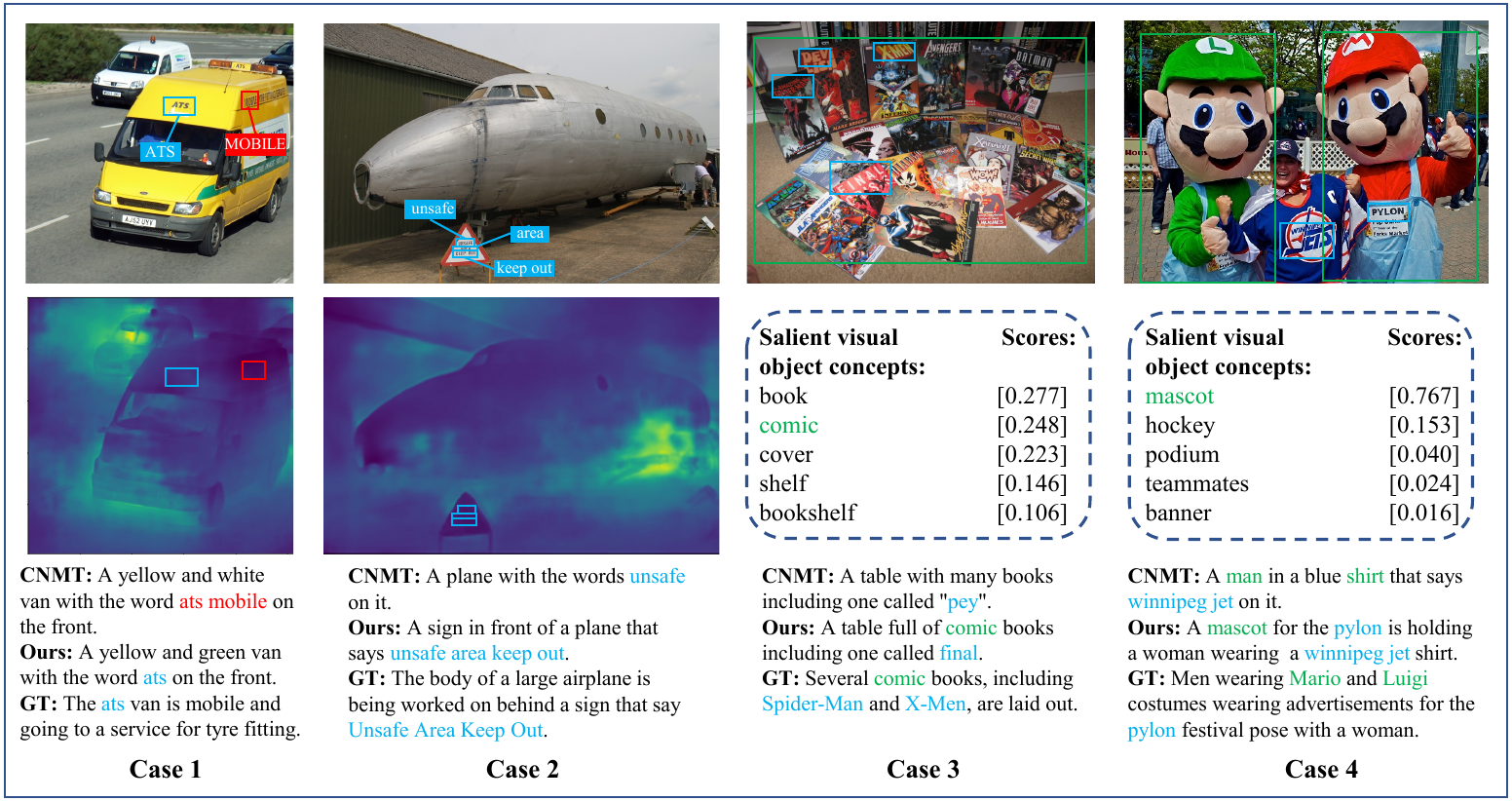}
   \caption{Example of captions generated by strong baseline CNMT \cite{wang2021confidence}, DEVICE, and GT (ground truth). \textcolor{red}{Red} indicates inappropriate sence text generated by CNMT. The depth maps and visual object concepts are displayed in the middle. The score of visual object concepts measures how similar these concepts are to the image. \textcolor[rgb]{0,0.76,0.98}{Blue} indicates scene text and \textcolor[rgb]{0,0.69,0.23}{Green} indicates visual object concepts.}
   \label{fig:case}
\end{figure*}

\subsection{Ablation Studies}
To analyze the respective effects of each module on the model performance, we conduct ablation experiments, as shown in \cref{tab3}. Comparing Line 1 (Rosetta only) and Line 2, CIDEr-D is boosted from $101.5$ to $105.3$. It is evident that more affluent and more accurate OCR tokens positively impact the model performance. While the improvement on BLEU-4 is relatively moderate, we think one reason may be that BLEU-4 treats all matched words equally, but CIDEr-D pays more attention to OCR tokens. Line $2$ and $3$ prove that simply concatenating depth values with two-dimensional space coordinates boosts the performance substantially, which increases BLEU-4 by $0.7$ and CIDEr-D by $4.5$, respectively. Comparing Line $3$ and $4$, we find that updating the appearance features of OCR tokens under the supervision of depth information boosts all metrics. Line $5$ indicates based on Line $2$ directly input semantic embeddings of visual concepts into the Multimodal Transformer. The introduction of visual concepts significantly improves the BLEU-4 and CIDEr-D. The significant improvements between Line $4$ and $7$ demonstrate the effectiveness of visual object concepts' semantic information, which boosts BLEU-4 by $1.3$ and CIDEr-D by $2.0$, respectively. Comparing Line 5 and 6, the introduction of SgAM reduces the iterations for the model to converge to the optimum, and improves the model's performance on CIDEr-D by $2.0$, respectively. Experimental data of Line $8$ demonstrates the robustness of DEVICE to the quality of OCR tokens. More remarkably, the total increase ratio of CIDEr-D is $11.8$ and BLEU-4 is $2.7$. This demonstrates that all modules work together very efficiently.

We also evaluate the impacts of the number $K$ of visual object concepts on the TextCaps validation (cf. \cref{tab5}). The results indicate that some salient visual object concepts may be missed when $K$ is less than~5. Meanwhile, irrelevant visual concepts may be retrieved when $K$ is larger than 7. Therefore, we set $K$ to $5$.

We show some cases which are generated by CNMT \cite{wang2021confidence} and DEVICE on the TextCaps validation set in \cref{fig:case},  where ``GT'' means ground truth. Case 1 and Case 2 demonstrate the effectiveness of visual entities' 3D spatial relationship modeling ability. Case 3 and 4 illustrate the influences of effectively utilize visual object concepts by SgAM. In Case 1, CNMT mistakenly takes ``ats'' and ``mobile'' as tokens in the same plane and connects them incorrectly. By introducing 3D geometric relationship, our DEVICE generates accurate scene text. In Case 2, CNMT incompletely represents the text in the sign and confuses the positional relationship between the airplane and the text on the sign. DEVICE enhances the correlation of scene text on the warning sign and clarifies the spatial relationship between visual entities by introducing 3D geometric relationships. We consider that the integration of depth-aware methods into tasks such as OCR recognition is capable of enhancing the accuracy of text chunk identification. For Case 3, CNMT only generates relatively coarse-grained words such as ``many books''. In contrast, with the help of semantic information of visual object concepts and SgAM, DEVICE is capable of generating more accurate words ``comic books'', like human. In the last case, unlike CNMT, which ignores the visual object ``mascots'', DEVICE generates a more comprehensive description, enabling rational use of scene text ``pylon''. The aforementioned cases demonstrate the effectiveness of DEVICE well. We consider the methodology of incorporating visual concepts can be extrapolated to the Human-Centric Image Captioning task \cite{yang2022human}, as a more comprehensive understanding of the scene facilitates the model in generating human-centric captions.

\begin{figure}[t]
  \centering
   \includegraphics[width=1.0\linewidth]{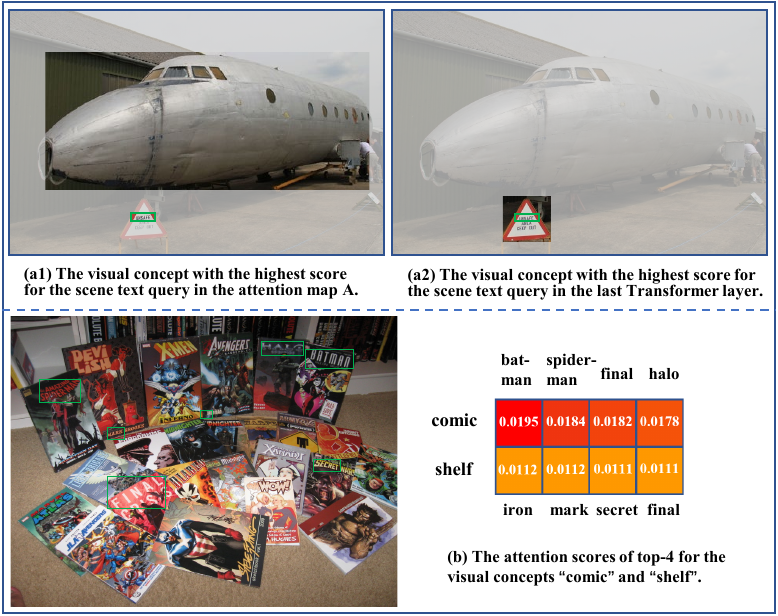}
   \caption{Visualization of the modules DeFUM and SgAM. For the DeFUM we select the bounding box with the best score as the image region that the model mostly focuses on when using scene text as queries. For the SgAM we list the OCR tokens with top-4 attention scores in the Semantic Attention.}
   \label{fig:visualization}
\end{figure}
\subsection{Qualitative Analysis}

However, in Case 4, we can see the gap between DEVICE and human, we consider one possible reason is that humans have common knowledge. Although adopting CLIP \cite{radford2021learning} to extract the visual object concept ``mascot'' in Case 4 can also be regarded as a way of implicitly using external knowledge, this knowledge is not fine-grained enough compared to ``Mario'' and ``Luigi''.

Moreover, we show the visualizations of the modules DeFUM and SgAM, cf. \cref{fig:visualization}. In Sub-figure (a), when using scene text ``unsafe'' as the query, we observe that the attention region shifted from the airplane (cf. Sub-figure (a1)) to the sign containing the text (cf. Sub-figure (a2)). We attribute this to the relatively smaller relative depth value between the scene text ``unsafe'' and the object ``sign'' compared to the relative depth value between the scene text ``unsafe'' and the object ``aircraft''. With the help of the depth-aware Self-Attention in DeFUM, DEVICE is capable of leveraging depth information to model the relationships between visual entities. In Sub-figure (b), the visual concept ``comic'' is highly relevant (bright red)
to the scene text ``batman'' and ``spiderman''. In contrast, the visual concept ``shelf'' exhibits a lower correlation (orange-red) with these OCR tokens, as there is not a significant semantic association between them. This indicates that the SgAM module can assist the interaction between visual entities and their associated scene text.

\begin{figure}[t]
  \centering
   \includegraphics[width=1.0\linewidth]{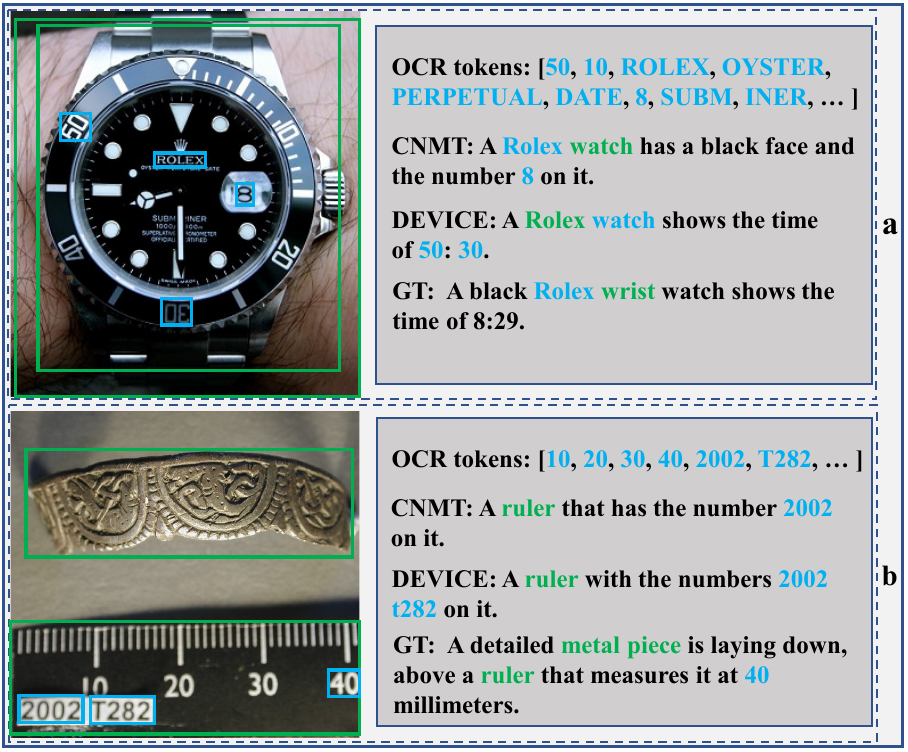}
   \caption{Negative samples of CNMT and DEVICE on the TextCaps validation set. The objects are inside the \textcolor[rgb]{0,0.69,0.23}{Green} bounding boxes, the OCR tokens are inside the \textcolor[rgb]{0,0.76,0.98}{Blue} bounding boxes. GT represents the Ground Truth.}
   \label{fig:case2}
\end{figure}

\subsection{Limitation Analysis}
We present negative examples generated by our model on the TextCaps validation dataset in \cref{fig:case2}. In Case a, while the strong baseline CNMT merely transcribes the characters on the dial, our model strives to extract higher-level information, such as time, but incorrectly interprets the watch's time. We consider the reason is that the ability of DEVICE to model the relationship between objects and scene text is still limited. Besides, it is difficult for models to read the time on the dial without numbers. In Case b, both CNMT and DEVICE fail to accurately capture the concept of "measuring length," indicating a general deficiency in current models' abilities to perceive intent within scenes.

\section{Conclusion}

In this paper, we propose a Depth and Visual Concepts Aware Transformer (DEVICE) for OCR-based image captinong, which is capable of generating accurate and comprehensive captions for given images. We introduce depth information and design a depth-enhanced feature updating module to improve OCR appearance features, aiding in 3D geometric relationship construction. This approach benefits tasks like OCR recognition. Meanwhile, we introduce the semantic information of salient visual object concepts and propose a semantic-guided alignment module to interact visual concepts with aligned scene text, which improves the integrity of captions. More remarkably, our DEVICE achieves state-of-the-art results on the TextCaps test set. However, certain scenarios challenge our model's interpretation of human knowledge, potentially overlooking high-level semantics. Comparing results with human annotations suggests that integrating explicit external knowledge could mitigate this issue.

\bibliographystyle{elsarticle-num}
\bibliography{egbib.bib} 



\end{document}